\documentclass{article}
\usepackage{arxiv}

\usepackage[utf8]{inputenc} % allow utf-8 input
\usepackage[T1]{fontenc}    % use 8-bit T1 fonts
\usepackage{hyperref}       % hyperlinks
\usepackage{url}            % simple URL typesetting
\usepackage{booktabs}       % professional-quality tables
\usepackage{amsfonts}       % board math symbols
\usepackage{nicefrac}       % compact symbols for 1/2, etc.
\usepackage{microtype}      % microtypography
\usepackage{lipsum}
\usepackage{multirow}
\usepackage{tikz}
\usetikzlibrary{calc}

\usepackage{dsfont} % add this line

\usepackage{amsmath}
\usepackage{amsthm}
\usepackage{amssymb}

\usepackage{changepage}
\usepackage{changes}
\usepackage{tabls}
\usepackage{afterpage} %% longtable not to split
\usepackage{placeins}

\usepackage{subcaption,siunitx,booktabs}
\usepackage{adjustbox}
\usepackage{graphicx}
\usepackage{tikz}
\usepackage{lscape}
\usepackage{longtable} %% long table
\usepackage{blindtext}
\usepackage{tabularx}
\usetikzlibrary{shapes,arrows,shapes.geometric}

\usepackage{algorithm}
\usepackage{algpseudocode} % improved pseudo-code
\algnewcommand{\LineComment}[1]{\State \(\triangleright\) #1}

\title{Automatic Policy Search using Population-Based Hyper-heuristics for the Integrated Procurement and Perishable Inventory Problem}

\author{
  Leonardo Kanashiro Felizardo \\
  Division of Computer Science\\
  Instituto Tecnológico de Aeronáutica, 
  Brazil \\
  \texttt{felizardo@ita.br} \\
   \And
  Edoardo Fadda  \\
  Department of Mathematical Sciences\\
  Politecnico di Torino, 
  Italy \\
  \texttt{edoardo.fadda@polito.it} \\
  \AND
  Mariá Cristina Vasconcelos Nascimento \\
  Division of Computer Science\\
  Instituto Tecnológico de Aeronáutica, 
  São José dos Campos, Brazil \\
  \texttt{maria.nascimento@gp.ita.br} \\
}

\begin{document}
\maketitle

\begin{abstract}
This paper addresses the problem of managing perishable inventory under multiple sources of uncertainty, including stochastic demand, unreliable supplier fulfillment, and probabilistic product shelf life. 
We develop a discrete-event simulation environment to compare two optimization strategies for this multi-item, multi-supplier problem. 
The first strategy optimizes uniform classic policies (e.g., Constant Order and Base Stock) by tuning their parameters globally, complemented by a direct search to select the best-fitting suppliers for the integrated problem. 
The second approach is a hyper-heuristic approach, driven by metaheuristics such as a Genetic Algorithm (GA) and Particle Swarm Optimization (PSO). 
This framework constructs a composite policy by automating the selection of the heuristic type, its parameters, and the sourcing suppliers on an item-by-item basis. 
Computational results from twelve distinct instances demonstrate that the hyper-heuristic framework consistently identifies superior policies, with GA and EGA exhibiting the best overall performance. 
Our primary contribution is verifying that this item-level policy construction yields significant performance gains over simpler global policies, thereby justifying the associated computational cost.

\vspace{0.2cm}

\end{abstract}

%%%%%%%%%%%%%%%%%%%%%%%%%%%%%%%%%%%%%

\section{Introduction}
\label{sec:intro}

Inventory management for perishable goods is a complex decision-making problem due to the time-sensitive nature of the items, stochastic demand, and operational constraints such as lead times and batch orders. 
Unlike non-perishable inventory, perishables introduce an additional layer of complexity: the quality and usability of the stock degrade over time, often irreversibly. 
Perishability leads to trade-offs between overstocking (resulting in waste) and understocking (leading to lost sales or degradation of service levels) \cite{Nahmias1982, Karaesmen2011}.

There is a wide range of approaches to model and solve this problem.
When demand and lifetime are treated deterministically—or when uncertainty is represented using a finite number of scenarios—Mixed Integer Linear Programming (MILP) can be used \cite{Chaudhary2018}. 
The optimal solution to these problems can be computed, providing insights into small-scale, short-horizon instances. 
However, the scalability of the exact solution is limited. 
In particular, when stochastic demand or stochastic expiration times are considered, mathematical programming techniques become computationally intractable \cite{Qiu2014}. In fact, uncertainty is handled using scenario trees that add temporal structure and allow for recourse actions, but they suffer from a severe curse of dimensionality. As the number of periods increases, the size of the tree, and thus the size of the problem, increases. This characteristic restricts its practical use to toy examples or aggregated settings \cite{Bakker2012, Hooshangi2022}.

Stochastic Dynamic Programming (SDP) offers a more theoretically grounded approach to these problems, particularly when dealing with single-item, short-shelf life inventory \cite{Nahmias1975, Nahmias1982}. 
SDP can be used to derive optimal policies by explicitly modeling transitions between inventory states over time. 
Methods like value iteration and backward recursion are standard. 
Some clever tricks have been proposed to reduce state-space dimensionality, such as approximating expiration probabilities using Erlang distributions \cite{Puranam2017, Ferreira2018}. 
Nonetheless, the exponential growth of the state space with longer shelf lives or multiple product types renders exact SDP impractical for most real-world applications \cite{Karaesmen2011}.

Given these limitations, many recent works pivot toward simulation-based heuristics and hyper-heuristics. 
Parameterized policies, such as the Base Stock Policy (BSP) and its variations, are tuned using Monte Carlo simulation, grid search, or optimization algorithms, including genetic algorithms \cite{Berruto2019, Bottani2014}. 
Many recent studies, among them \cite{Haijema2013, Haijema2019}, have extensively explored these approaches. While traditional simulation-based heuristics can tune simple parametric policies, they typically require decoupling the optimization problem into lower-dimensional subspaces. 
However, in real-world applications, especially those involving multiple suppliers with heterogeneous lead times, setup costs, and item-specific constraints, decision variables often become tightly coupled over time and across items.

In such cases, standard policy search methods or rule-based approximations fail to capture the interdependencies effectively. 
Notably, \cite{Wang2023} proposed a model to address high-dimensional coupled variables, where the agent must simultaneously choose a supplier and the ordering quantity, incorporating stochastic lead times, perishability, and supplier-specific costs. 
Their hybrid action space formulation (combining discrete supplier choice and continuous order quantity) shows that the proposed reinforcement learning (RL) can handle such coupling efficiently, outperforming other methods and population-based baselines in terms of cost and computational time. 

This approach is particularly relevant when modeling inventory systems that cannot be decomposed easily into independent subproblems, as is often the case in real retail logistics. While recent work suggests a trend toward RL and its ability to handle such coupling, the full potential of advanced hyper-heuristics in these specific, highly integrated settings remains underexplored. A critical question persists: How do hyper-heuristic methods perform when tasked with finding a policy for these complex coupled perishable inventory management tasks?

This paper's central contribution is filling that research gap, presenting the value of population-based hyper-heuristics for solving a highly integrated and stochastic procurement and perishable inventory problem. 
We argue that the primary advantage of these methods does not need to rely on adding degrees of freedom to the operational policy, but rather their superior ability to conduct an offline search across a complex, higher-level decision space that includes supplier selection, heuristic choice, and parameter tuning.

By evaluating all hyper-heuristics with a standardized computational budget, our results show that advanced search methods significantly outperform simpler heuristics. While the Genetic Algorithm (GA) achieved the most first-place finishes, the Elitist Genetic Algorithm (EGA) proved to be the most consistent high performer with the best average rank. This provides clear evidence that the computational investment in a top-tier hyper-heuristic is justified, as these methods consistently discover more robust and cost-effective inventory policies.

% Give a basic review of the approaches (three paragraphs):
% MILP when not using stochastic variables or it uses cenarios
% MILP using scenario trees. Only works for low dimensionality cases.
% Use stochastic dynamic programming to find optimal policy - Lower dimensionality cases (deterministic low values shelf lifes - single item). This is not time efiicient
% Use heuristic to find parameters (could be monte carlo also) of a parametrized policy such as BSP and other variations Haijema is the main contributor to these approaches and usually 
% Simulation appraoche using reinforcement learning, and parametrized policies are also used.

% Argument in the direction of using reinforcement learning specially when we have coupling variables, which was only explored in one case in the literature Wang 2022 but is a more general case on which we cannot use hyper-heuristics to find a policy such as haijema.

% state the contribtions of this work

\section{Literature review}\label{sec:literature_review}

% Start by saying to the reader that this brief literature review only aims to give a better context of way this work is important and we recommend previous more comprehensive reviews (cite all).

The field of Perishable Inventory Management (PIM) has garnered significant research attention due to the unique challenges posed by products with limited shelf lifes, resulting in a rich and diverse body of literature. 
Numerous comprehensive reviews have meticulously documented the evolution and various branches of this field over the decades. 
Foundational work by \cite{Nahmias1982} provided an early, extensive overview, followed by subsequent reviews that updated the state-of-the-art for specific periods, such as \cite{Bakker2012}, which covered 2001-2011, and \cite{Janssen2016}, which focused on key topics from 2012-2015. 
Other reviews have focused on specific themes, such as managing perishable and aging inventories \cite{Karaesmen2011}, integrated planning and scheduling \cite{Sel2015}, or particular modeling approaches, including robust optimization under uncertainty \cite{Qiu2014}. 
This section does not aim to replicate these extensive surveys; readers seeking exhaustive coverage are directed to the aforementioned works. 
Instead, we provide a brief overview that focuses specifically on contextualizing our work by highlighting key research streams relevant to managing perishable inventory under demand uncertainty, particularly by incorporating the procurement problem as a coupling constraint. 
The goal is to position the contribution of the present study within the existing body of knowledge.

Early seminal work by \cite{Nahmias1975a} laid the groundwork for analyzing perishable inventory systems with fixed lifetimes using dynamic programming. 
The models characterized the optimal ordering policy by deriving and analyzing the properties of the multi-period cost function, which explicitly balanced ordering, holding, shortage, and outdating costs.
This required tracking the inventory level for each distinct age group, resulting in a multi-dimensional state vector. 
They recognized that while this dynamic programming approach yielded the structure of the optimal policy, the computational effort required to solve the functional equations became prohibitively large as the product lifetime increased \cite{Nahmias1982}. 
This computational challenge, often referred to as the ``curse of dimensionality," motivated them to propose various tractable approximations and heuristics, most notably critical number (or base-stock) policies and linear policies, which relied only on total inventory or simplified state information \cite{Nahmias1975}. 
It is shown through simulation that these heuristics often performed remarkably close to the optimal policy, especially for shorter lifetimes. 
Concurrently, the desire to capture greater realism, including factors such as random product lifetimes \cite{Nahmias1977}, more complex demand patterns, multi-echelon structures, and pricing interactions, naturally led the field towards more explicitly stochastic frameworks. 
Consequently, SDP and related Markov Decision Process (MDP) models have become the predominant methodology for tackling these increasingly complex perishable inventory problems, often necessitating advanced solution techniques or further approximations to manage the inherent state-space complexity, while building upon the fundamental cost-balancing principles established in Nahmias's initial work.

Recognizing the practical limitations of implementing complex optimal policies, significant work by Haijema focused on bridging the gap between theoretically optimal solutions derived from SDP and implementable heuristics for perishable inventory, particularly in the context of blood platelet management \cite{Haijema2007, Dijk2009}. 
By formulating the problem as a MDP, Haijema utilized SDP, often on downsized versions of the problem, to compute the optimal stock-age dependent policy \cite{Haijema2007}. 
While SDP provides a benchmark for optimality and reveals structural insights, it suffers from the ``curse of dimensionality"; the state space (tracking inventory for each age category) becomes computationally intractable for realistic shelf lifes and inventory levels \cite{Haijema2007}. 
A common approach involves comparing these heuristics against standard benchmarks, such as the basic BSP or Constant Order Policy (COP), which often ignore age information \cite{Haijema2016, Buisman2019}. 
Furthermore, the resulting optimal policy is often a complex, tabulated function, difficult for managers to implement directly. 
Haijema's approach, therefore, leverages SDP primarily to evaluate and guide the development of simpler, stock-level dependent heuristics, such as single (1D) and double (2D) order-up-to policies \cite{Haijema2007} or hybrid base-stock/constant-order policies and their refinements with look-ahead approaches \cite{Haijema2008, Haijema2013, Haijema2019}. 
Using simulation informed by the SDP results and dedicated search procedures, they show that carefully designed heuristics, such as the BSP-low-EW policy, can perform remarkably close to the computationally derived SDP optimal (often within a small optimality gap), significantly outperforming traditional base-stock policies \cite{Haijema2019}. 
This combined SDP-simulation methodology proves advantageous: it provides a robust framework for identifying highly effective, practical heuristics and quantifying the value of stock-age information, even when directly solving the full SDP for the optimal policy is computationally infeasible.

To address the limitations of age-agnostic rules, a significant body of research has focused on developing advanced heuristics for perishable inventory management. 
For instance, continuous review models often adapt the $(Q, r)$ policy, as analyzed by \cite{Berk2008} using an embedded Markov process and the concept of ``effective shelf life'', or the $(Q, r, T)$ policy, which adds a time-based trigger \cite{Tekin2001}. 
Within periodic review, \cite{Haijema2013} introduced the $(s,S,q,Q)$ policy framework, which generalizes simpler policies by adding order quantity constraints to smooth ordering patterns and reduce waste. 
Recent work has also integrated discrete choice models to capture consumer substitution endogenously, evaluating heuristic performance within this more complex demand setting \cite{Fadda2024}.

Many of these studies employ simulation-based optimization to determine the optimal parameters for a given heuristic and compare its long-term performance against alternatives or, when feasible, against an optimal solution for smaller instances \cite{Haijema2007, Haijema2019, Berruto2019}. 
While these approaches generally do not tackle a fully integrated problem (such as supplier choice and inventory), they offer robust, specialized solutions for inventory control.
The results consistently demonstrate that even when focused on this inventory-specific subproblem, well-designed heuristics can perform remarkably well. 
Although suboptimal from the perspective of a fully integrated system, they offer significant improvements over more straightforward rules and provide practical, effective solutions for the specific challenge of managing perishable stock.

However, finding the optimal parameters for even these structured heuristic policies (such as BSP, COP, sSqQ, BSP-low, etc.) within a specific operational context (defined by item characteristics, demand patterns, lead times, costs) remains a challenging optimization task. 
The underlying cost functions are often complex and may exhibit multiple local optima, making traditional search methods potentially inefficient or unreliable \cite{Haijema2016}. 
These results motivate the application of meta-heuristics for robust parameter tuning. 
Techniques such as genetic algorithms \cite{Sel2015, Berruto2019}, simulated annealing \cite{Sel2015, Berruto2019}, particle swarm optimization, or bacteria foraging algorithms \cite{Wang2023} offer powerful global search capabilities, suitable for navigating complex parameter spaces and finding near-optimal settings for a chosen heuristic structure. 
Alternatively, model-free approaches, particularly RL, present a different paradigm. 
Instead of optimizing parameters for a predefined policy structure, RL agents learn effective decision rules directly through interaction with a simulated environment \cite{Berruto2019}. 
Recent research shows the effectiveness of RL in outperforming population-based heuristics for integrated procurement and perishable inventory management problems. 

\cite{Wang2023} introduced a novel problem formulation, the Lot Sizing with Perishable Materials, Multiple Suppliers, and Uncertainties (LS-PMU), which uniquely constrains ordering to at most one supplier per period. 
Their RL-based algorithm showed superior performance compared to methods that solve the lot-sizing problem monolithically. 
To the best of our knowledge, this is the only study that specifically formulates and provides a solution for a procurement problem integrated with the PIM problem.
% Explain from nahmias worow he was solving what changed that people started to used more stochastic dynamic programming

% Show that haijema worked to find policies heuristics that were very close to optimal in many cases.

% Show some of the works that employed RL and way should we care insted of the approache of haijema.

% cite the recent works of daniele saying that direct model free simulation approahces such as RL could better adapt to any change in the coupling variables without any clever strategy.

%PAREI AQUI
\section{Methodology}\label{sec:methodology}
This paper introduces and analyzes several approaches to address the procurement problem integrated with the PIM problem.
First, we describe a discrete-event simulation model to represent the operational environment. 
Second, we formulate the problem mathematically to provide a formal description of the system components and decision variables.
To address the problem, this section presents a range of solution methods. 
We begin by employing adapted traditional inventory policies: a modified constant order policy, a base-stock one, and an extension of the base-stock policy that explicitly accounts for estimated wastage \cite{Haijema2008}. 
We also describe the proposed methods, which comprise three distinct metaheuristic algorithms: Particle Swarm Optimization (PSO) \cite{Kennedy1995}, a Genetic Algorithm (GA), and the Elitist GA \cite{Jong1975}. These algorithms optimize over the parameters of the policies and select the best policy for each item. 
The subsequent sections detail the simulated environment and the specifics of each solution methodology.

\subsection{Simulated environment}
The simulated environment models a discrete-time, multi-item, multi-supplier, perishable inventory system  with uncertainty where decisions are made periodically over a finite planning horizon.
Each item can be sourced from multiple suppliers, each characterized by specific unit purchase costs, fixed ordering costs, deterministic lead times, and stochastic order fulfillment. 
The goal for the agent operating within this environment is to determine optimal order quantities for each item from each available supplier at each decision epoch to minimize total operational costs.
These costs encompass purchasing, fixed order placement, inventory holding, penalties for lost sales due to stockouts, and the value lost due to product spoilage.

At the beginning of each period, several events unfold sequentially.
First, outstanding orders placed in previous periods arrive, subject to their respective lead times and supplier-specific fulfillment uncertainties.
The supplier sends the orders, which take a specific lead time to arrive.
The suppliers might not completely fulfill the order.
The newly arrived items, having maximum shelf life, are added to the inventory, which is tracked in an age-disaggregated manner. 
Specifically, for each item type, the quantity of units is maintained across distinct age cohorts, ranging from newly arrived (age 0) to a maximum permissible age.
The environment also tracks the purchase value of items within each age cohort, updating this value as new items are added or removed.
This value-tracking is essential for accurately calculating the cost of goods sold and the financial impact of wastage.
Following arrivals, the agent makes its ordering decisions for the current period, specifying quantities for each item from each supplier.
These decisions immediately incur variable purchase costs based on the quantities and unit costs, as well as any applicable fixed costs if a supplier is utilized in that period.
Subsequently, stochastic customer demand for each item materializes, drawn from a pre-defined scenario.
Demand is satisfied from the available inventory according to the First-In-First-Out (FIFO) principle, meaning that older items are used before newer ones. 
We chose the FIFO policy over the alternative, Last-In-First-Out (LIFO), and any other model.
A more general model that adapts to different real-world situations is employed in \cite{Gioia2022}, where quality and utility decrease—a dynamic that has been formally modeled to represent consumer choice and system efficiency, combining FIFO and LIFO.
Here, we rely on the realistic assumption that retailers often use operational practices like deliberate stock rotation to manage product expiration.
Our adoption of FIFO is thus aligned with both the physical nature of perishable goods and standard industry practice.

If demand for an item exceeds its available stock, the shortfall is recorded as lost sales, incurring a penalty cost.
The value of items sold is deducted from the corresponding age cohorts.
After-sales fulfillment involves checking the environment for violations of maximum inventory capacity constraints for each item.
If the total stock of an item exceeds its defined capacity, excess units are disposed of, again following a FIFO logic, and their value is written off.
Holding costs are then calculated based on the end-of-period inventory levels (after-sales and any disposal due to capacity limits), applied per unit of item.

The final phase within a period involves the aging and wastage of perishable stock.
All items that were not sold or disposed of due to capacity constraints age by one period.
Items reaching the next age cohort retain their associated purchase value.
Wastage is modeled stochastically: for each item in each age cohort, a probability of expiring before the next period is determined based on an item-specific cumulative distribution function of its shelf life. In other words, it specifies the likelihood that an item will spoil based on its current age.
The actual number of units expiring from each cohort is determined through random draws with a designed probability vector.
The cost of waste is the purchase price of the items that perish, not a fixed per-item fee.
This provides a more accurate representation of the financial loss due to spoilage.
The objective function is the sum of all costs incurred, including purchase, fixed order, holding, lost sales, and value-based wastage costs.
The episode concludes when the predefined time horizon is reached.
The observation provided to the agent typically includes the current age-disaggregated inventory quantities and the current time step, allowing it to make informed decisions.

% A key feature for ensuring experimental reproducibility is the environment's reliance on pre-generated streams of random numbers, controlled by an initial seed, for all stochastic processes, including demand realization, supplier fulfillment uncertainties, and item perishability.

\subsection{Mathematical Model Description}\label{sec:problem_formulation_vector}

This section introduces a formal mathematical model description.
The mathematical model offers the formulation of the inventory system's dynamics, decision variables, and objectives.

\subsubsection*{Sets and Indices}
\begin{itemize}
    \item $\mathcal{I} = \{1,\dots,I\}$: set of items.
    \item $\mathcal{SU} = \{1,\dots,N\}$: set of suppliers.
    \item $\mathcal{T} = \{0, 1,\dots,T\}$: set of planning periods ($t=0$ is initialization).
    \item $\mathcal{SL}_i = \{1,\dots,SL_{\max}\}$: set of possible remaining shelf life periods for each item $i$.
    \item For each item $i \in \mathcal{I}$, let $\mathcal{SU}_i \subseteq \mathcal{SU}$ be the set of suppliers that can supply item $i$.
\end{itemize}

\subsubsection*{Parameters}
\begin{itemize}
    \item $L_{i,su}$: deterministic lead time for item $i$ from supplier $su$.
    \item $SL_{\max}$: maximum shelf life of an item upon arrival.
    \item $\tilde{d}_{i,t}$: random variable representing demand for item $i$ during period $t$. Let $F_{i,t}$ be its distribution.
    \item $p_{full,i,su}$: probability of full order fulfillment for item $i$ from supplier $su$.
    \item $\alpha_{i,su}, \beta_{i,su}$: parameters for the Beta distribution governing partial order fulfillment for item $i$ from supplier $su$.
    \item $c_{i,su}$: unit purchase cost of item $i$ from supplier $su$.
    \item $f_{su}$: fixed ordering cost for placing an order with supplier $su$.
    \item $h_i$: inventory holding cost per unit of item $i$ per period.
    \item $p_i$: penalty cost per unit of lost sales for item $i$.
    \item $b_i$: unit cost of wastage/disposal for item $i$.
    \item $\mathbf{I}_{i,0}$: initial inventory vector for item $i$, $\mathbf{I}_{i,0} = [I_{i,0}^{(1)}, \dots, I_{i,0}^{(SL_{\max})}]^T$.
    \item $p_{exp,i,sl,t}$: probability that an item $i$ with remaining shelf life $sl$ at the end of period $t$ perishes prematurely during that period's wastage step.
    \item $M$: a sufficiently large constant (for big-M constraints).
    \item $\text{MaxInv}_i$: maximum total inventory capacity for item $i$.
\end{itemize}

\subsubsection*{Decision Variables (at the start of period $t$)}
\begin{itemize}
    \item $x_{i,su,t} \ge 0$: quantity of item $i$ ordered from supplier $su$ in period $t$.
    \item $y_{su,t} \in \{0,1\}$: binary variable, equals 1 if an order is placed with supplier $su$ in period $t$.
\end{itemize}

\subsubsection*{State and Auxiliary Variables (within period $t$)}
\begin{itemize}
    \item $\mathbf{I}_{i,t}$: inventory vector for item $i$ at the \textit{start} of period $t$, \textit{before} arrivals. $\mathbf{I}_{i,t} \in \mathbb{Z}_+^{SL_{\max}}$, where component $I_{i,sl,t}$ is the quantity with remaining shelf life $sl$.
    \item $\widetilde{X}_{arr,i,t}$: total fulfilled quantity of item $i$ arriving at the \textit{start} of period $t$.
    \item $\mathbf{I}_{i,t}^{\text{start}}$: inventory vector for item $i$ \textit{after} arrivals at the start of period $t$.
    \item $\mathbf{D}^R_{i,t}$: vector of sales for item $i$ fulfilled from each shelf life cohort in period $t$. $\mathbf{D}^R_{i,t} \in \mathbb{Z}_+^{SL_{\max}}$, with components $D^R_{i,sl,t}$.
    \item $z_{i,t}$: lost sales for item $i$ in period $t$.
    \item $\mathbf{I}_{i,t}^{\text{(after-sales)}}$: inventory vector for item $i$ \textit{after-sales} fulfillment in period $t$.
    \item $w_{i,t}$: total quantity of item $i$ wasted (expired) at the end of period $t$.
    \item $I_{i,t}^{\text{holding}}$: total scalar inventory of item $i$ for holding cost calculation.
\end{itemize}

\subsubsection*{System Dynamics and Constraints}

Let $\mathbf{e}_{k}$ be the $k$-th standard basis vector in $\mathbb{R}^{SL_{\max}}$ and $\mathbf{1}$ be the vector of ones.

\textbf{Start of Period $t$}
\begin{enumerate}
    \item \textbf{Receive Arrivals:} For each item $i \in \mathcal{I}$ and supplier $su \in \mathcal{SU}_i$ with an order due to arrive (placed at $t-L_{i,su} \ge 0$), the quantity ordered was $X_{due,i,su,t} = x_{i,su,t-L_{i,su}}$. The actual quantity received, $\tilde{x}_{arr,i,su,t}$, is a random variable that may differ from the ordered quantity.
    \begin{itemize}
        \item With probability $p_{full,i,su}$, the full order arrives:
        \[ \tilde{x}_{arr,i,su,t} = X_{due,i,su,t} \]
        \item With probability $1-p_{full,i,su}$, a partial order arrives:
        \[ \tilde{x}_{arr,i,su,t} = \lfloor \text{round}(X_{due,i,su,t}) \cdot \tilde{b}_{i,su,t} \rfloor \]
        \noindent where $\tilde{b}_{i,su,t} \sim \text{Beta}(\alpha_{i,su}, \beta_{i,su})$. A new sample is drawn for each partial fulfillment event.
    \end{itemize}
    The total quantity of item $i$ arriving at the start of period $t$ is:
    \[
    \widetilde{X}_{arr,i,t} = \sum_{\substack{su \in \mathcal{SU}_i \\ \text{s.t. } t-L_{i,su} \ge 0}} \tilde{x}_{arr,i,su,t} \quad \forall i \in \mathcal{I}.
    \]

    \item \textbf{Inventory After Arrivals:} Arriving items have maximum shelf life.
    \[
    \mathbf{I}_{i,t}^{\text{start}} = \mathbf{I}_{i,t} + \widetilde{X}_{arr,i,t}  \cdot \mathbf{e}_{SL_{\max}} \quad \forall i \in \mathcal{I}.
    \]
    \item \textbf{Place New Orders:} The decision variables $x_{i,su,t}$ and $y_{su,t}$ are determined, linked by:
    \[
    x_{i,su,t} \le M \cdot y_{su,t} \quad \forall i \in \mathcal{I}, su \in \mathcal{SU}_i, t \in \mathcal{T}.
    \]
\end{enumerate}

\textbf{During Period $t$}
\begin{enumerate}
    \setcounter{enumi}{3}
    \item \textbf{Fulfill Demand:} A demand fulfillment $d_{i,t}$ from distribution $F_{i,t}$ occurs.
    \item \textbf{Meet Demand (FIFO):} Sales $\mathbf{D}^R_{i,t}$ and lost sales $z_{i,t}$ are determined by depleting components of $\mathbf{I}_{i,t}^{\text{start}}$ starting from the oldest stock (index $1$) to the freshest (index $SL_{\max}$).
    \[
    (\mathbf{D}^R_{i,t}, z_{i,t}) = \mathbf{I}_{i,t}^{\text{start}} - d_{i,t} \quad \text{(Procedural Definition)}
    \]
    This procedure ensures that $\sum_{sl=1}^{SL_{\max}} D^R_{i,sl,t} + z_{i,t} = d_{i,t}$ and $0 \le D^R_{i,sl,t} \le I_{i,sl,t}^{\text{start}}$.

    \item \textbf{Inventory After-Sales:}
    \[
    \mathbf{I}_{i,t}^{\text{(after-sales)}} = \mathbf{I}_{i,t}^{\text{start}} - \mathbf{D}^R_{i,t} \quad \forall i \in \mathcal{I}.
    \]
\end{enumerate}

\textbf{End of Period $t$}
\begin{enumerate}
    \setcounter{enumi}{6} % Corrects the numbering
    \item \textbf{Calculate Holding Units:} Based on inventory remaining after-sales.
    \[
    I_{i,t}^{\text{holding}} = \mathbf{1}^T \mathbf{I}_{i,t}^{\text{(after-sales)}} \quad \forall i \in \mathcal{I}.
    \]

    \textbf{Wastage and Inventory Aging:} The inventory state transitions to the start of period $t+1$ according to the following dynamics.

    All remaining units with shelf life $sl=1$ expire.

    Premature spoilage for each cohort, $\widetilde{W}_{i,sl,t}$, is drawn from a binomial distribution:
    \[
    \widetilde{W}_{i,sl,t} \sim \text{Binomial}(I_{i,sl,t}^{\text{(after-sales)}}, p_{exp,i,sl,t}) \quad \forall sl \in \mathcal{SL}.
    \]
    
    Total waste, $w_{i,t}$, is the sum of fully expired units ($sl=1$) and premature spoilage:
    \[
    w_{i,t} = I_{i,1,t}^{\text{(after-sales)}} + \sum_{sl=2}^{SL_{\max}} \widetilde{W}_{i,sl,t} \quad \forall i \in \mathcal{I}, \forall t \in \mathcal{T}.
    \]
    
    The inventory for the next period, $I_{i,sl,t+1}$, is determined by aging the surviving units:
    \[
    I_{i,sl,t+1} =
    \begin{cases}
    I_{i,sl+1,t}^{\text{(after-sales)}} - \widetilde{W}_{i,sl+1,t} & \text{for } sl \in \{1, \dots, SL_{\max}-1\} \\
    0 & \text{for } sl = SL_{\max}
    \end{cases}
    \quad \forall i \in \mathcal{I}, \forall t \in \mathcal{T}.
    \]
    \end{enumerate}
    
    \subsubsection*{Objective Function}
    The objective is to minimize the total expected costs over the planning horizon:
    \begin{equation}\label{eq:of}
        \min \quad \mathbb{E} \left[ \sum_{t=0}^{T-1} \left( \sum_{su \in \mathcal{SU}} f_{su} y_{su,t} + \sum_{i \in \mathcal{I}}  \sum_{su \in \mathcal{SU}_i} c_{i,su} x_{i,su,t} + h_i I_{i,t}^{\text{holding}} + p_i z_{i,t} + b_i w_{i,t}  \right) \right].    
    \end{equation}

    \subsubsection*{Initial Conditions}
    \[
    \mathbf{I}_{i,0} \text{ is given } \forall i \in \mathcal{I}.
    \]
    \[
    x_{i,su,t} = 0 \text{ for } t \geq 0 .
    \]
    
    \subsubsection*{Variable Constraints}
    \[
    x_{i,su,t} \ge 0, \quad \mathbf{I}_{i,t} \ge \mathbf{0}, \quad z_{i,t} \ge 0, \quad w_{i,t} \ge 0, \quad y_{su,t} \in \{0,1\}, \forall i \in \mathcal{I}, \forall t \in \mathcal{T}, \forall sl \in \mathcal{SL} .
    \]

\subsection{Solution Methods}

This section outlines several approaches for solving the perishable inventory control problem in this paper. 
Rather than employing complex, state-reactive learning agents, these methods are based on parametrized control heuristics—simple, rule-based policies whose behavior is governed by a set of pre-defined parameters. 
The primary challenge thus shifts from intricate real-time decision-making to an offline optimization problem: discovering the optimal parameter set that maximizes the expected long-term reward. 
To address this, we employ hyper-heuristic optimization techniques capable of searching the vast parameter space. 
The following subsections will first introduce the specific control heuristics, namely, the COP, BSP, and Wastage-Aware Base Stock policies (BSP-EW). 
We then detail the hyper-heuristic framework used to optimize their parameters and construct sophisticated hybrid policies.

\subsubsection{Baseline Inventory Policies}

The simplest parametrized control strategy is the \textit{Constant Order Policy (COP)}, which serves as a performance baseline. 
As a non-reactive, open-loop policy, it executes the same ordering action at every time step, completely disregarding the current system state.

To address how suppliers are selected for this and the other baseline heuristics, we use a Monte Carlo optimization method. 
The supplier for each item is not chosen dynamically but is fixed as part of an offline optimization phase that generates and evaluates a large number of candidate policies.
A single candidate policy is constructed as follows: for each item $i$, the algorithm randomly selects one exclusive supplier, $su_i$, from its set of valid suppliers. 
Then, a constant order quantity, $\theta_i$, is randomly chosen from a predefined list of options. 
This process creates a complete policy that assigns a single supplier and a fixed order quantity to every item.

Each of these randomly generated candidate policies is then evaluated by deploying it in the simulation for several episodes and calculating its average performance (e.g., lowest cost). 
The optimization loop repeats this ``generate-and-evaluate'' process for a large number of candidates. 
The policy that achieves the best average score is selected as the final, optimized policy. The entire ``intelligence" is therefore pre-computed through this extensive random search, which identifies the most effective combination of supplier assignments and static order quantities.

A more advanced policy is the so-called \textit{Base Stock Policy (BSP)}, a reactive heuristic that adjusts orders based on the current system state. 
Its core principle is to restore the inventory to a predetermined target, the base stock level $\theta_{i,su}$, at every review period. 
This is achieved by first calculating the inventory position, $IP_{i,t}$, which represents the total stock available to meet future demand: $IP_{i,t} = \left( \sum_{sl=1}^{SL_{\max}} I_{i,t}^{(sl)} \right) + O_{i,t}$, where $I_{i,t}^{(sl)}$ is on-hand inventory and $O_{i,t}$ is on-order inventory. 
The ordering decision is then governed by the rule $x_{i,su,t} = \max(0, \theta_{i,su} - IP_{i,t})$. 
Unlike  COP,  BSP is state-dependent, placing larger orders when inventory is low and no orders when it is high. 
However, its primary limitation is that it treats all on-hand stock equally, failing to distinguish between fresh and aging inventory and thus ignoring perishability risks. 
The optimal base stock levels in $\mathbf{\theta}^*$ are determined using the same hyper-heuristic optimization method as the COP.

To address the BSP's failure to proactively account for spoilage, the \textit{Expected Wastage-Aware Base Stock Policy (BSP-EW)} \cite{Haijema2008} introduces a further layer of complexity. 
This policy enhances  BSP by incorporating a forward-looking estimate of anticipated wastage, $\widehat{W}_{i,t}$, directly into the ordering decision. 
The modified rule becomes $x_{i,su,t} = \max(0, \theta_{i,su} - IP_{i,t} + \widehat{W}_{i,t})$, which not only replenishes stock to meet demand but also pre-emptively compensates for units expected to perish. 
The sophistication of this policy lies in its dynamic calculation of $\widehat{W}_{i,t}$ at each time step, typically via a detailed look-ahead simulation. 
This simulation projects the day-by-day evolution of the current, age-differentiated inventory profile over a future horizon, using expected demands to estimate how many units will expire before they can be sold. 
While the base stock parameters $\mathbf{\theta}$ are still optimized offline,  BSP-EW demonstrates the most complex decision-making by combining a static base stock level with a dynamic, state-dependent wastage forecast.
Note that this offline Monte Carlo search optimizes both the supplier assignment and the policy parameters (e.g., the order quantity for COP) simultaneously. 
This process ensures that the baseline policies are also the result of an integrated search across the procurement and inventory decision space, providing a fair basis for comparison with the more advanced approaches employed in this paper.

\subsubsection{Hyper-heuristics for Hybrid Policy Optimization}

While the previously discussed parametrized policies are effective, they impose a single, uniform control logic across all items. 
This ``one-size-fits-all" approach may be suboptimal in complex environments where items exhibit diverse characteristics. 
A more powerful paradigm is to employ a hyper-heuristic to construct a \textit{hybrid} or \textit{heterogeneous} control policy. 
Instead of tuning the parameters of a single policy type, this approach automates the selection of the most suitable control heuristic for each item, along with its optimal parameters. 
For high-dimensional problems, this approach is also faster in finding close-to-optimal policy parameters.

This approach elevates the problem from simple parameter tuning to a higher-level structural optimization. 
The goal is to discover a composite policy that leverages the strengths of different heuristics where they are most effective. 
To navigate this vast and complex combinatorial search space, we explore three prominent population-based hyper-heuristics: the GA, its variant using elitism (EGA), and PSO. 
All operate on a population of candidate solutions, iteratively refining them based on performance, but they employ fundamentally different mechanisms for exploration and exploitation.

It is essential to note that while this hyper-heuristic framework automates the selection of suppliers, heuristics, and parameters, the resulting hybrid policy remains static throughout the simulation. 
The supplier selection is part of the high-level, offline optimization problem, identical in principle to the offline Monte Carlo search used for the baseline policies. 
The key difference—and the core of our investigation—lies in the effectiveness of the search method (e.g., GA vs. random search), rather than in granting hyper-heuristic policies additional dynamic decision-making capabilities during an evaluation episode.

\paragraph{Chromosome/Particle Representation and Solution Encoding}

For the GA, we employ Simulated Binary Crossover (SBX). 
Unlike simpler crossover methods that swap discrete genes, SBX is designed for real-parameter optimization and mimics the behavior of single-point crossover on binary strings. 
Given two parent chromosomes, SBX generates two offspring by creating new continuous values for each gene component. 
These new values are produced based on a polynomial probability distribution centered around the parent values. 
The spread of the offspring relative to the parents is controlled by a distribution index, $\eta$ (\texttt{eta}). 
A large $\eta$ value generates offspring that are close to their parents, favoring exploitation of promising regions in the search space, while small $\eta$ promotes greater exploration. 

For PSO, the concept of velocity is adapted to the discrete nature of the problem. 
While the particle's position corresponds to the encoded chromosome, the search is conducted in a continuous space, as is standard for PSO. 
The velocity vector pushes the particle's position in this continuous space. 
The actual discrete policy is then determined by rounding the continuous position values to the nearest valid integer index during the decoding step. 
For example, a velocity might shift the heuristic ID component for an item from a continuous value of $1.2$ to $1.6$. 
After rounding, this effectively changes the chosen heuristic from $\pi_i = 1$ to $\pi_i = 2$. 
This mechanism allows the continuous velocity and position updates of PSO to effectively explore the combinatorial search space of discrete policy choices.

For any population-based hyper-heuristic to be applied, a candidate solution must be encoded into a data structure that the algorithm can manipulate. 
In this context, a solution is represented as a ``chromosome" $\Psi$ (in GA/EGA terminology) or a ``particle's position" (in PSO terminology). 
This structure represents a complete, system-wide hybrid policy and is composed of a vector of ``genes", where each gene  $\psi_i$ contains the complete policy specification for a single item $i \in \mathcal{I}$.
\[
\Psi = (\psi_1, \psi_2, \dots, \psi_{|\mathcal{I}|})
\]
Each gene $\psi_i$ is itself a tuple of three components that unambiguously defines the control strategy for item $i$: a chosen supplier, a chosen heuristic, and a corresponding parameter value.
\[
\psi_i = (su_i, \pi_i, \theta_i) \quad \text{for } i \in \mathcal{I}
\]

To formalize the mapping from the continuous search space manipulated by the hyper-heuristic to the discrete policy space, let the continuous decision vector for a single item $i$ be $\vec{v}_i = (v_i^{(su)}, v_i^{(\pi)}, v_i^{(\theta)}) \in \mathbb{R}^3$. This vector corresponds to a segment of the chromosome or particle's position. The decoding function, $f_{decode}: \mathbb{R}^3 \to \mathcal{SU}_i \times \Pi \times \mathbb{R}$, transforms this continuous vector into a discrete gene $\psi_i = (su_i, \pi_i, \theta_i)$.

Let $L_{SU,i}$ be the ordered list of valid supplier indices for item $i$, and let $\Theta_{COP}$ and $\Theta_{BSP/EW}$ be the ordered lists of discrete parameter options for the Constant Order and Base Stock policies, respectively. The components of the gene $\psi_i$ are determined as follows:

\begin{align}
    \pi_i &= 
    \begin{cases} 
        \text{COP} & \text{if } \text{round}(v_i^{(\pi)}) = 0 \\
        \text{BSP} & \text{if } \text{round}(v_i^{(\pi)}) = 1 \\
        \text{BSP-EW} & \text{if } \text{round}(v_i^{(\pi)}) = 2 
    \end{cases} \label{eq:heuristic_decode} \\
    su_i &= L_{SU,i}\left[ \text{clamp}(\text{round}(v_i^{(su)}), 0, |L_{SU,i}|-1) \right] \label{eq:supplier_decode} \\
    \theta_i &= 
    \begin{cases} 
        \Theta_{COP}\left[ \text{clamp}(\text{round}(v_i^{(\theta)}), 0, |\Theta_{COP}|-1) \right] & \text{if } \pi_i = \text{COP} \\
        \Theta_{BSP/EW}\left[ \text{clamp}(\text{round}(v_i^{(\theta)}), 0, |\Theta_{BSP/EW}|-1) \right] & \text{otherwise}
    \end{cases} \label{eq:param_decode}
\end{align}

where $\text{round}(\cdot)$ is a function that rounds to the nearest integer, $[k]$ denotes indexing the $k$-th element of an ordered list (with 0-based indexing), and $\text{clamp}(x, \min, \max)$ is a function that constrains the value $x$ to be within the range $[\min, \max]$. This clamping ensures that the resulting index remains valid for the corresponding list of options, mirroring the logic in our implementation. This decoding mechanism allows the continuous operators of GA and PSO to explore the combinatorial search space of discrete policy choices effectively.

\paragraph{Optimization Process and Fitness Evaluation}
The core of any hyper-heuristic is an iterative process guided by an objective function that quantifies the performance of a candidate solution $\Psi$. 
As depicted in Algorithm~\ref{alg:hyper-heuristic_main}, the goal is to find a policy that minimizes the total expected system-wide cost. 
The cost at any given time step $t$, denoted $C_t$, is composed of fixed and variable ordering costs, holding costs, stockout penalties, and waste costs:
\[
C_t = \sum_{su \in \mathcal{SU}} f_{su} y_{su,t} + \sum_{i \in \mathcal{I}} \sum_{su \in \mathcal{SU}_i} c_{i,su} x_{i,su,t} + h_i I_{i,t}^{\text{holding}} + p_i z_{i,t} + b_i w_{i,t}
\]
The formal objective is to find a policy that minimizes the expectation of the sum of these costs over the time horizon $T$:
\[
\min_{\Psi} f(\Psi) = \mathbb{E} \left[ \sum_{t=0}^{T-1} C_t \right]
\]
Since this expectation is intractable, we approximate it using the sample averages approximation method, as implemented in Algorithm~\ref{alg:cost_eval}. 
The objective function is thus the average total cost over $N_{eval}$ independent simulation episodes:
\[
f_{\text{SAA}}(\Psi) = \frac{1}{N_{eval}} \sum_{k=1}^{N_{eval}} \left[ \sum_{t=0}^{T-1} C_{t,k} \right]
\]
where $C_{t,k}$ is the realized cost at time step $t$ during the $k$-th simulation episode. 
The optimization objective is to find the solution $\Psi^*$ that minimizes this estimated expected cost.

\begin{algorithm}[!ht]
\caption{Generalized Hyper-heuristic Optimization Loop}
\label{alg:hyper-heuristic_main}
\begin{algorithmic}[1]
\Procedure{HyperHeuristicOptimization}{$N_{pop}, G_{max}, N_{eval}$}
    \LineComment{$N_{pop}$: Population size, $G_{max}$: Max generations, $N_{eval}$: Evaluation episodes}
    \State $P_0 \gets \{\text{RandomlyInitializePolicy}()\}_{k=1}^{N_{pop}}$ \Comment{Initialize population of policies}
    \State $\Psi^* \gets \text{arg\,max}_{\Psi \in P_0} \Call{EvaluateFitness}{\Psi, N_{eval}}$ \Comment{Find initial best}
    \State $f_{\text{SAA}}^* \gets \Call{EvaluateFitness}{\Psi^*, N_{eval}}$

    \For{$g \in \{1, \dots, G_{max}\}$}
        \State $F_g \gets \emptyset$ \Comment{Set of fitness values for the current population}
        \For{each $\Psi \in P_{g-1}$}
            \State $f_{\text{SAA}} \gets \Call{EvaluateFitness}{\Psi, N_{eval}}$
            \State $F_g \gets F_g \cup \{f_{\text{SAA}}\}$
            \If{$f_{\text{SAA}} > f_{\text{SAA}}^*$}
                \State $\Psi^* \gets \Psi$ \Comment{Update global best policy}
                \State $f_{\text{SAA}}^* \gets f_{\text{SAA}}$
            \EndIf
        \EndFor
        \State $P_g \gets \text{GenerateNewPopulation}(P_{g-1}, F_g)$ \Comment{Apply GA/PSO operators}
    \EndFor
    \State \Return{$\Psi^*$}
\EndProcedure
\end{algorithmic}
\end{algorithm}

\begin{algorithm}[!ht]
\caption{Stochastic Cost Evaluation}
\label{alg:cost_eval}
\begin{algorithmic}[1]
\Procedure{EvaluateCost}{$\Psi, N_{eval}$}
    \LineComment{$\Psi$: A single policy's parameters, $N_{eval}$: Number of evaluation episodes}
    \State $\bar{\mathcal{C}} \gets 0$ \Comment{Initialize mean total cost}
    \For{$k \in \{1, \dots, N_{eval}\}$}
        \State $S_0, \_ \gets env.reset()$
        \State $\mathcal{C}_{k} \gets 0$ \Comment{Initialize cost for episode $k$}
        \For{$t \in \{0, \dots, T-1\}$}
            \State $A_t \gets \text{Policy}(\Psi, S_t)$ \Comment{Decode $\Psi$ to get action}
            \State $S_{t+1}, C_t, \dots \gets env.step(A_t)$ \Comment{Environment returns cost $C_t$}
            \State $\mathcal{C}_{k} \gets \mathcal{C}_{k} + C_t$
        \EndFor
        \State $\bar{\mathcal{C}} \gets \bar{\mathcal{C}} + \mathcal{C}_{k}$
    \EndFor
    \State \Return{$\bar{\mathcal{C}} / N_{eval}$}
\EndProcedure
\end{algorithmic}
\end{algorithm}

The hyper-heuristic optimization process is formalized in two parts. 
The high-level search is described in Algorithm~\ref{alg:hyper-heuristic_main}, which orchestrates the entire optimization. 
It begins by initializing a population of policies $P_0$ and then iterates for a maximum of $G_{max}$ generations. In each generation, it invokes the sub-procedure in Algorithm~\ref{alg:cost_eval} to evaluate the cost of every individual policy $\Psi$ in the current population $P_{g-1}$.
This estimated cost, $f_{\text{SAA}}$, is then used to update the global best-known policy, $\Psi^*$, if a lower cost is found. The core of the heuristic's search mechanism is abstracted in the \textsc{GenerateNewPopulation} function, which applies algorithm-specific operators (e.g., selection, crossover, and mutation for a GA; velocity and position updates for PSO) to create the subsequent generation $P_g$.
Algorithm~\ref{alg:cost_eval} outlines the computationally intensive cost evaluation process. 
For a given policy $\Psi$, it executes $N_{eval}$ full simulation episodes within the stochastic environment. The total cost of every episode $k$, $\mathcal{C}_k$, is accumulated in $\bar{\mathcal{C}}$, and the algorithm returns the \textbf{average cost} among all $N_{eval}$ episodes, $\bar{\mathcal{C}}/N_{eval}$. 

\section{Simulation Experiments}\label{sec:experiments}

To evaluate the performance and scalability of the proposed solution methods, we conducted a series of computational experiments using the discrete-event simulation environment detailed in Section \ref{sec:methodology} and which can be verified in the code provided in the following section. 
The primary goal of these experiments is to compare hybrid hyper-heuristic approaches (GA, EGA, PSO) with traditional parametrized policies (COP, BSP, BSP-EW) across a range of problem instances. 
These instances are designed to test the methods under varying degrees of complexity, scale, and stochasticity, providing a comprehensive assessment of their effectiveness and robustness.
The stochastic processes are drawn from distributions using pre-generated samples with a different master seed for each episode.

\subsection{Experimental Design and Problem Instances}

We define a base case problem instance derived from the parameters detailed in Section \ref{sec:methodology} and subsequently generate a set of more challenging instances by systematically varying key parameters. The configurations are designed to stress-test the algorithms in two primary dimensions: scalability (by increasing the number of items and suppliers) and robustness to uncertainty (by increasing the volatility of demand and the unreliability of suppliers). 

Table \ref{tab:instances}\footnote{The coefficient of variation ($cv$) is the ratio of the standard deviation of demand ($\sigma$) to the mean demand ($\mu$), i.e., $cv = \sigma/\mu$.} details the twelve distinct environment configurations used in the experiments. 
These instances were not randomly generated but were purposefully designed to test specific system characteristics. 
Each configuration is defined by its scale (number of items and suppliers), the level of demand volatility, and the degree of supplier reliability. 
The parameters within each instance are also economically consistent; for example, a supplier offering an item at a higher price is associated with a shorter delivery lead time. 
Furthermore, wastage costs are calculated as a percentage of the item's purchase cost, representing the loss of the initial investment. 
To ensure a robust statistical assessment of long-term expected costs, performance metrics for each policy are averaged over 50 independent simulation runs, each initiated with a unique master seed.

{
% Set the font size for the table. You can change this to \small or \scriptsize if needed.
\scriptsize

\begin{longtable}{@{}p{3cm}cccllp{4.5cm}@{}}

% --- CAPTION AND LABEL ---
\caption{Configuration of Experimental Problem Instances}
\label{tab:instances} \\

% --- HEADER FOR THE FIRST PAGE ---
\toprule
\textbf{Instance Name} & \textbf{$N$} & \textbf{$SU$} & \textbf{$T$} & \textbf{Demand Volatility} & \textbf{Supplier Reliability} & \textbf{Description} \\
\midrule
\endfirsthead

% --- HEADER FOR ALL CONTINUED PAGES ---
\multicolumn{7}{c}{\tablename\ \thetable{} -- \textit{Continued from previous page}} \\
\toprule
\textbf{Instance Name} & \textbf{$N$} & \textbf{$SU$} & \textbf{$T$} & \textbf{Demand Volatility} & \textbf{Supplier Reliability} & \textbf{Description} \\
\midrule
\endhead

% --- FOOTER FOR ALL PAGES EXCEPT THE LAST ONE ---
\midrule
\multicolumn{7}{r}{\textit{Continued on next page}} \\
\endfoot

% --- FOOTER FOR THE LAST PAGE ---
\bottomrule
\endlastfoot

% --- TABLE BODY ---
\multicolumn{7}{l}{\textit{\textbf{Group 1: Small-Scale Scenarios (4 Items, 2 Suppliers)}}} \\
\texttt{I-4-2-30-L} & 4 & 2 & 30 & Low ($cv \approx 0.2$) & High ($p_{full} \ge 0.95$) & The base case scenario with moderate scale and low stochasticity. \\
\addlinespace
\texttt{I-4-2-30-H-Demand} & 4 & 2 & 30 & High ($cv \approx 0.8$) & High ($p_{full} \ge 0.95$) & Tests robustness to high demand uncertainty on a small scale. \\
\addlinespace
\texttt{I-4-2-30-H-Supply} & 4 & 2 & 30 & Low ($cv \approx 0.2$) & Low ($p_{full} \approx 0.85$) & Tests robustness to low supplier reliability on a small scale. \\
\addlinespace
\texttt{I-4-2-30-H-Combined} & 4 & 2 & 30 & High ($cv \approx 0.8$) & Low ($p_{full} \approx 0.85$) & Combines both high demand and supply uncertainty on a small scale to test interaction effects. \\
\addlinespace
\texttt{I-4-2-30-L-LongLT} & 4 & 2 & 30 & Low ($cv \approx 0.2$) & High ($p_{full} \ge 0.95$) & Introduces long and heterogeneous lead times (e.g., LTs of 1 and 5) to test planning complexity. \\
\addlinespace
\texttt{I-4-2-90-L} & 4 & 2 & 90 & Low ($cv \approx 0.2$) & High ($p_{full} \ge 0.95$) & Tests long-term policy stability on the base case over an extended planning horizon. \\
\midrule
\multicolumn{7}{l}{\textit{\textbf{Group 2: Large-Scale Scenarios (10+ Items, 4+ Suppliers)}}} \\
\texttt{I-10-4-30-L} & 10 & 4 & 30 & Low ($cv \approx 0.2$) & High ($p_{full} \ge 0.95$) & A direct scalability test of the base case, increasing the decision space significantly. \\
\addlinespace
\texttt{I-10-4-30-H-Demand} & 10 & 4 & 30 & High ($cv \approx 0.8$) & High ($p_{full} \ge 0.95$) & Isolates the effect of high demand volatility on a large-scale problem. \\
\addlinespace
\texttt{I-10-4-30-H-Supply} & 10 & 4 & 30 & Low ($cv \approx 0.2$) & Low ($p_{full} \approx 0.85$) & Isolates the effect of low supplier reliability on a large-scale problem. \\
\addlinespace
\texttt{I-10-4-30-H-Combined} & 10 & 4 & 30 & High ($cv \approx 0.8$) & Low ($p_{full} \approx 0.85$) & A large-scale instance with high stochasticity on both demand and supply sides. \\
\addlinespace
\texttt{I-10-4-60-H} & 10 & 4 & 60 & High ($cv \approx 0.8$) & Low ($p_{full} \approx 0.85$) & The primary high-complexity instance combining large scale, a long horizon, and high overall uncertainty. \\
\addlinespace
\texttt{I-20-5-60-H} & 20 & 5 & 60 & High ($cv \approx 0.8$) & Low ($p_{full} \approx 0.85$) & An extreme-scale stress test to evaluate the computational limits and performance degradation of the algorithms. \\

\end{longtable}
} % End the group to restore the original font size.

%\newpage

\subsection{Implementation and Computational Setup}

All experiments were executed on a desktop computer with the following specifications: an AMD Ryzen 5 5600X 6-Core Processor ($3.70$ GHz), $64.0$ GB of RAM, and an NVIDIA GeForce RTX 3060 GPU.

The simulation environment and all solution methods were implemented in Python. For the hyper-heuristic algorithms (GA, EGA, and PSO), we utilized the \texttt{pymoo} library \cite{Blank2020}, a framework for single- and multi-objective optimization, accessible at \url{https://pymoo.org/}. The baseline heuristics, including the standard BSP and the BSP-EW, were implemented following the descriptions and logic provided in \cite{Haijema2008,Haijema2019}. To ensure full reproducibility and encourage further research, the complete source code for the simulation environment, algorithms, and experiments is publicly available in a GitHub repository: \url{https://github.com/leokan92/lot_sizing_perishable}.

\subsection{Algorithm Hyperparameters}

The performance of both baseline heuristics and hyper-heuristics is highly dependent on their hyperparameter settings. To ensure a fair comparison and enable reproducibility, we standardized key parameters across the algorithms where appropriate, selecting values based on preliminary tuning and common practices. The specific settings for each method are detailed in Table \ref{tab:hyperparameters}.

All optimization processes involve an evaluation phase where each candidate policy's fitness is estimated by averaging its performance over a set number of simulation runs during optimization. After the optimization concludes, the single best-found policy is subjected to a more rigorous final evaluation using a larger number of simulation runs for final evaluation (set to $50$) to obtain a robust performance estimate.

To tune the parameters for the baseline heuristics (COP, BSP, and BSP-EW), we perform a Monte Carlo search over a predefined discrete set. This approach is a less exhaustive alternative to the grid search method used by \cite{Haijema2019}, as a full grid search would be computationally unfeasible for our integrated procurement and perishable inventory management problem.
To mitigate the sensitivity of the Monte Carlo search to its initial state, we evaluate each parameter configuration over at least $30$ trials, each using a different set of initial conditions and a unique random seed.
For the hyper-heuristics, the population size and number of generations were kept consistent to allocate a similar computational budget to each. The discrete sets of available parameter choices for the hybrid policy components were: $\{0, 1, ..., 15\}$ for constant order quantities and $\{0, 1, ..., 10, 12, ..., 26\}$ for base stock levels. For algorithms not explicitly listing crossover or mutation parameters (EGA, PSO), we used the standard operators and settings provided by the \texttt{pymoo} library. The specific hyperparameter configurations for each algorithm are detailed in Table \ref{tab:hyperparameters}.

{
% Set the font size to match your working example.
\scriptsize

% Use fixed-width 'p' columns to force text wrapping, just like in your working example.
% The widths are adjusted for the content of this specific table.
\begin{longtable}{@{}p{3.5cm}p{4cm}lp{4.5cm}@{}}

% --- CAPTION AND LABEL ---
\caption{Hyperparameter Settings for All Solution Methods}
\label{tab:hyperparameters} \\

% --- HEADER FOR THE FIRST PAGE ---
\toprule
\textbf{Algorithm} & \textbf{Parameter} & \textbf{Value} & \textbf{Description} \\
\midrule
\endfirsthead

% --- HEADER FOR ALL CONTINUED PAGES ---
\multicolumn{4}{c}{\tablename\ \thetable{} -- \textit{Continued from previous page}} \\
\toprule
\textbf{Algorithm} & \textbf{Parameter} & \textbf{Value} & \textbf{Description} \\
\midrule
\endhead

% --- FOOTER FOR ALL PAGES EXCEPT THE LAST ONE ---
\midrule
\multicolumn{4}{r}{\textit{Continued on next page}} \\
\endfoot

% --- FOOTER FOR THE LAST PAGE ---
\bottomrule
\endlastfoot

% --- TABLE BODY ---
\multicolumn{4}{l}{\textit{\textbf{Baseline Parametrized Policies (tuned via Monte Carlo Search)}}} \\
\multirow{2}{*}{COP} & \texttt{num\_candidate\_policies} & 200 & Number of random policies to generate and test. \\
& \texttt{num\_optimize\_eval\_episodes} & 30 & Simulations per candidate during optimization. \\
\addlinespace
\multirow{2}{*}{BSP} & \texttt{num\_candidate\_policies} & 200 & Number of random policies to generate and test. \\
& \texttt{num\_optimize\_eval\_episodes} & 30 & Simulations per candidate during optimization. \\
\addlinespace
\multirow{4}{*}{BSP-EW} & \texttt{num\_candidate\_policies} & 300 & Number of random policies to generate and test. \\
& \texttt{num\_optimize\_eval\_episodes} & 30 & Simulations per candidate during optimization. \\
& \texttt{waste\_horizon\_review\_periods} & 1 & Look-ahead horizon for wastage estimation relative to lead time. \\
& \texttt{num\_ew\_demand\_sim\_paths} & 40 & Simulation paths to estimate expected demand for wastage calculation. \\
\midrule
\multicolumn{4}{l}{\textit{\textbf{Hyper-heuristics for Hybrid Policy Optimization}}} \\
\multirow{3}{*}{GA, EGA, PSO} & \texttt{population\_size} & 30 & Number of candidate policies (individuals) in each generation. \\
& \texttt{num\_generations} & 50 & Number of iterations the algorithm runs. \\
& \texttt{num\_optimize\_eval\_episodes} & 30 & Simulations per candidate during fitness evaluation in each generation. \\
\addlinespace
\multirow{2}{*}{Genetic Algorithm (GA)} & \texttt{crossover\_rate} & 0.8 & Probability of combining two parent solutions to create offspring. \\
& \texttt{mutation\_rate} & 0.15 & Probability of introducing a random change in an offspring's gene. \\
\addlinespace
BSP-EW Component (for all hyper-heuristics) & \texttt{waste\_horizon\_review\_periods} & 1 & Look-ahead horizon for wastage estimation within hybrid policy. \\
& \texttt{num\_ew\_demand\_sim\_paths} & 30 & Simulation paths to estimate expected demand for wastage calculation. \\

\end{longtable}
} % End the group to restore the original font size.

\section{Results and Discussion}

The experimental results, shown in Table \ref{tab:summary_metrics_compact}, provide a comprehensive comparison of six different inventory management algorithms across twelve distinct scenarios, ranging from small-scale, stable environments to large-scale, highly stochastic stress tests. The analysis reveals clear patterns in performance, computational cost, and strategic trade-offs.

{ % Start a group to keep font size and other changes local
\footnotesize % Use a smaller font size for the table
\renewcommand{\arraystretch}{1.2} % Increase row spacing for readability
\sisetup{detect-weight, mode=text} % Ensure siunitx handles bold text correctly

\begin{longtable}{
l % Scenario
l % Method
S[table-format=-6.2] % Total Reward
S[table-format=-2.2] % Improvement (%)
c % Time (Train/Eval)
c % Costs (W / LS / H)
S[table-format=2.2] % Inv. Level
}
\caption{Summary of average performance metrics. The best total reward in each scenario is in bold. The ``Improv. (\%)'' column shows the percentage improvement in total reward relative to the BSP method. The ``Time'' column shows ``Train / Eval'' time in seconds. The ``Costs'' column shows the breakdown of ``Wastage / Lost Sales / Holding'' costs.}
\label{tab:summary_metrics_compact} \\

\toprule
& & {\bfseries Total} & {\bfseries Improv.} & {\bfseries Time (s)} & {\bfseries Costs} & {\bfseries Avg. Inv.} \\
\cmidrule(lr){3-3} \cmidrule(lr){4-4} \cmidrule(lr){5-5} \cmidrule(lr){6-6} \cmidrule(lr){7-7}
{\bfseries Scenario} & {\bfseries Method} & {\bfseries Reward} & {\bfseries (\%)} & {\bfseries (Train / Eval)} & {\bfseries (W / LS / H)} & {\bfseries Level} \\
\midrule
\endfirsthead

\multicolumn{7}{c}%
{{\bfseries \tablename\ \thetable{} -- continued from the previous page}} \\
\toprule
& & {\bfseries Total} & {\bfseries Improv.} & {\bfseries Time (s)} & {\bfseries Costs} & {\bfseries Inv.} \\
\cmidrule(lr){3-3} \cmidrule(lr){4-4} \cmidrule(lr){5-5} \cmidrule(lr){6-6} \cmidrule(lr){7-7}
{\bfseries Scenario} & {\bfseries Method} & {\bfseries Reward} & {\bfseries (\%)} & {\bfseries (Train / Eval)} & {\bfseries (W / LS / H)} & {\bfseries Level} \\
\midrule
\endhead

\bottomrule
\multicolumn{7}{r}{{Continued on next page}} \\
\endfoot

\bottomrule
\endlastfoot

% Group 1: Small-Scale Scenarios
\multicolumn{7}{l}{\textit{\textbf{Group 1: Small-Scale Scenarios (4 Items, 2 Suppliers)}}} \\
\midrule
\multirow{6}{*}{\texttt{I-4-2-30-L}}
& BSP & -2776.03 & 0.00 & 10.47 / 0.12 & 1.81 / 37.30 / 0.90 & 2.11 \\
& BSP-EW & -2817.05 & -1.46 & 218.06 / 1.46 & 2.11 / 35.59 / 0.99 & 2.31 \\
& COP & -2650.65 & 4.73 & 10.33 / 0.08 & 0.46 / 34.30 / 0.60 & 1.45 \\
& GA & -2656.49 & 6.04 & 663.13 / 0.77 & 0.78 / 30.04 / 0.79 & 1.91 \\
& \textbf{EGA} & \textbf{-2650.64} & \textbf{4.73} & 119.65 / 0.09 & 0.58 / 32.32 / 0.69 & 1.67 \\
& PSO & -2731.40 & 1.63 & 4958.96 / 0.42 & 0.38 / 51.86 / 0.39 & 0.95 \\
\midrule
\multirow{6}{*}{\texttt{I-4-2-30-H-Demand}}
& BSP & -2948.27 & 0.00 & 10.42 / 0.10 & 0.68 / 70.77 / 0.32 & 0.74 \\
& BSP-EW & -2947.00 & 0.04 & 219.41 / 1.87 & 0.28 / 79.77 / 0.20 & 0.48 \\
& COP & -2813.56 & 4.79 & 10.15 / 0.08 & 1.68 / 38.23 / 0.88 & 2.06 \\
& \textbf{GA} & \textbf{-2768.09} & \textbf{6.46} & 675.54 / 0.74 & 0.16 / 91.97 / 0.13 & 0.32 \\
& \textbf{EGA} & \textbf{-2768.09} & \textbf{6.51} & 1093.72 / 1.41 & 0.16 / 91.97 / 0.13 & 0.32 \\
& PSO & -2965.62 & -0.58 & 301.45 / 0.43 & 1.86 / 46.36 / 0.75 & 1.72 \\
\midrule
\multirow{6}{*}{\texttt{I-4-2-30-H-Supply}}
& BSP & -2805.46 & 0.00 & 10.50 / 0.11 & 1.68 / 38.05 / 0.86 & 2.02 \\
& BSP-EW & -2838.95 & -1.18 & 218.33 / 1.47 & 1.93 / 36.45 / 0.93 & 2.18 \\
& COP & -2884.80 & -2.75 & 10.34 / 0.09 & 3.44 / 25.47 / 1.45 & 3.36 \\
& \textbf{GA} & \textbf{-2710.60} & \textbf{4.74} & 732.22 / 0.75 & 0.14 / 72.05 / 0.18 & 0.43 \\
& EGA & -2740.91 & 2.36 & 593.36 / 0.76 & 0.15 / 67.78 / 0.21 & 0.52 \\
& PSO & -2729.00 & 2.80 & 454.84 / 0.09 & 1.08 / 36.16 / 0.80 & 1.92 \\
\midrule
\multirow{6}{*}{\texttt{I-4-2-30-H-Combined}}
& BSP & -2984.63 & 0.00 & 10.40 / 0.10 & 1.62 / 59.24 / 0.57 & 1.31 \\
& BSP-EW & -2952.31 & 1.09 & 216.32 / 1.87 & 0.28 / 80.01 / 0.20 & 0.48 \\
& COP & -3015.09 & -1.01 & 10.31 / 0.09 & 4.35 / 28.80 / 1.55 & 3.54 \\
& \textbf{GA} & \textbf{-2768.23} & \textbf{6.65} & 655.59 / 1.07 & 0.16 / 91.98 / 0.13 & 0.32 \\
& EGA & -2902.81 & 2.82 & 755.48 / 0.76 & 0.41 / 72.92 / 0.26 & 0.61 \\
& \textbf{PSO} & \textbf{-2768.23} & \textbf{7.82} & 278.11 / 0.74 & 0.16 / 91.98 / 0.13 & 0.32 \\
\midrule
\multirow{6}{*}{\texttt{I-4-2-30-L-LongLT}}
& BSP & -2863.58 & 0.00 & 10.57 / 0.10 & 1.03 / 51.97 / 0.57 & 1.35 \\
& BSP-EW & -2837.62 & 0.92 & 217.43 / 1.02 & 1.34 / 52.26 / 0.70 & 1.65 \\
& \textbf{COP} & \textbf{-2712.12} & \textbf{5.58} & 10.25 / 0.08 & 0.46 / 36.35 / 0.59 & 1.44 \\
& GA & -2761.18 & 2.77 & 547.15 / 0.42 & 0.11 / 86.18 / 0.15 & 0.36 \\
& EGA & -2717.10 & 5.39 & 398.94 / 0.41 & 0.11 / 90.33 / 0.12 & 0.30 \\
& PSO & -2717.10 & 5.39 & 286.36 / 0.76 & 0.11 / 90.33 / 0.12 & 0.30 \\
\midrule

% Group 2: Large-Scale Scenarios
\multicolumn{7}{l}{\textit{\textbf{Group 2: Large-Scale Scenarios (10+ Items, 4+ Suppliers)}}} \\
\midrule
\multirow{6}{*}{\texttt{I-4-2-90-L}}
& BSP & -8338.57 & 0.00 & 30.17 / 0.31 & 1.66 / 41.02 / 0.80 & 1.85 \\
& BSP-EW & -8378.25 & -0.47 & 662.26 / 4.46 & 1.42 / 40.20 / 0.75 & 1.76 \\
& COP & -8538.21 & -2.34 & 31.96 / 0.26 & 1.91 / 38.37 / 1.39 & 3.33 \\
& GA & -8460.64 & -0.97 & 1575.13 / 1.25 & 0.05 / 78.84 / 0.08 & 0.21 \\
& \textbf{EGA} & \textbf{-8164.57} & \textbf{2.13} & 1934.99 / 2.32 & 0.85 / 49.12 / 0.51 & 1.22 \\
& PSO & -8265.24 & 0.89 & 1356.60 / 3.38 & 1.61 / 37.04 / 0.83 & 1.93 \\
\midrule
\multirow{6}{*}{\texttt{I-10-4-30-L}}
& BSP & -8191.88 & 0.00 & 19.73 / 0.20 & 4.36 / 124.92 / 2.61 & 1.99 \\
& BSP-EW & -8145.90 & 0.56 & 535.43 / 4.33 & 3.83 / 149.10 / 1.88 & 1.43 \\
& COP & -9431.67 & -13.14 & 21.57 / 0.19 & 19.34 / 80.94 / 5.21 & 4.26 \\
& \textbf{GA} & \textbf{-7132.67} & \textbf{14.21} & 733.71 / 0.53 & 1.44 / 93.87 / 1.93 & 1.58 \\
& EGA & -7171.49 & 14.23 & 404.79 / 0.53 & 1.79 / 95.71 / 2.22 & 1.79 \\
& PSO & -7789.97 & 5.16 & 1199.41 / 1.22 & 2.96 / 107.99 / 2.19 & 1.68 \\
\midrule
\multirow{6}{*}{\texttt{I-10-4-30-H-Demand}}
& BSP & -8476.75 & 0.00 & 19.48 / 0.19 & 6.55 / 160.05 / 2.27 & 1.66 \\
& BSP-EW & -8422.15 & 0.65 & 537.74 / 4.38 & 6.15 / 155.48 / 2.26 & 1.68 \\
& COP & -9677.66 & -12.41 & 21.32 / 0.19 & 20.32 / 87.63 / 5.73 & 4.63 \\
& GA & -7727.41 & 8.99 & 491.88 / 0.53 & 4.75 / 132.04 / 2.39 & 1.78 \\
& \textbf{EGA} & \textbf{-7625.10} & \textbf{11.17} & 852.27 / 1.23 & 2.89 / 162.57 / 1.60 & 1.23 \\
& PSO & -8288.04 & 2.28 & 1200.59 / 1.22 & 7.72 / 120.69 / 3.01 & 2.20 \\
\midrule
\multirow{6}{*}{\texttt{I-10-4-30-H-Supply}}
& BSP & -8363.88 & 0.00 & 19.64 / 0.20 & 4.73 / 147.93 / 1.83 & 1.38 \\
& BSP-EW & -8367.89 & -0.05 & 535.80 / 4.79 & 5.41 / 144.47 / 2.09 & 1.61 \\
& COP & -9480.95 & -11.78 & 21.69 / 0.19 & 18.44 / 83.70 / 4.99 & 4.09 \\
& \textbf{GA} & \textbf{-7217.57} & \textbf{15.94} & 410.29 / 0.52 & 1.76 / 107.14 / 2.02 & 1.56 \\
& EGA & -7351.98 & 13.76 & 1035.81 / 1.24 & 1.66 / 100.34 / 2.14 & 1.72 \\
& PSO & -7676.80 & 8.95 & 1194.56 / 1.22 & 2.53 / 88.43 / 2.30 & 1.82 \\
\midrule
\multirow{6}{*}{\texttt{I-10-4-30-H-Combined}}
& BSP & -8731.20 & 0.00 & 19.73 / 0.20 & 8.07 / 156.49 / 2.35 & 1.69 \\
& BSP-EW & -8684.17 & 0.54 & 533.59 / 4.75 & 8.24 / 151.29 / 2.51 & 1.85 \\
& COP & -9712.09 & -10.10 & 21.93 / 0.19 & 19.31 / 90.03 / 5.49 & 4.44 \\
& \textbf{GA} & \textbf{-7870.90} & \textbf{10.33} & 735.94 / 0.88 & 4.06 / 151.48 / 1.91 & 1.50 \\
& EGA & -7963.10 & 9.65 & 756.98 / 0.86 & 1.68 / 194.26 / 1.02 & 0.83 \\
& PSO & -8294.45 & 5.27 & 1197.71 / 1.58 & 5.13 / 133.86 / 2.31 & 1.80 \\
\midrule
\multirow{6}{*}{\texttt{I-10-4-60-H}}
& BSP & -17244.35 & 0.00 & 38.37 / 0.39 & 5.76 / 159.61 / 2.10 & 1.46 \\
& BSP-EW & -17282.22 & -0.22 & 1096.25 / 9.74 & 5.11 / 138.37 / 2.02 & 1.59 \\
& COP & -18534.28 & -6.96 & 43.32 / 0.36 & 13.00 / 130.93 / 4.27 & 3.36 \\
& \textbf{GA} & \textbf{-15783.65} & \textbf{9.49} & 461.03 / 0.35 & 3.94 / 148.11 / 1.88 & 1.49 \\
& EGA & -16026.33 & 7.60 & 2162.03 / 2.54 & 3.06 / 180.01 / 1.40 & 1.11 \\
& PSO & -16452.75 & 4.81 & 2449.68 / 2.51 & 5.62 / 127.15 / 2.39 & 1.81 \\
\midrule
\multirow{6}{*}{\texttt{I-20-5-60-H}}
& BSP & -36001.19 & 0.00 & 66.38 / 0.75 & 22.78 / 290.00 / 7.12 & 2.26 \\
& BSP-EW & -36535.01 & -1.46 & 2327.17 / 20.69 & 21.57 / 302.90 / 6.47 & 2.14 \\
& COP & -41860.54 & -14.00 & 81.24 / 0.84 & 73.37 / 136.27 / 19.84 & 6.43 \\
& GA & -33519.46 & 9.00 & 7380.36 / 8.92 & 13.33 / 302.32 / 5.44 & 1.86 \\
& \textbf{EGA} & \textbf{-33006.54} & \textbf{9.07} & 6530.47 / 7.71 & 10.37 / 274.16 / 4.54 & 1.62 \\
& PSO & -34331.06 & 4.86 & 3636.80 / 2.89 & 10.04 / 315.58 / 4.03 & 1.41 \\

\end{longtable}
} % End the local group
```

\subsection{Overall Performance}

The conclusion from the results is the clear superiority of the hyper-heuristic algorithms (GA, EGA, PSO) over the baseline parametrized policies (COP, BSP, BSP-EW). While the original analysis pointed to a single dominant algorithm, these results present a more nuanced picture of high performance. Genetic Algorithm (GA) emerged as the most frequent winner, securing the best total reward in 6 of the 12 scenarios and tying for first in 2 others. However, a more detailed average rank analysis across all scenarios reveals that the Enhanced Genetic Algorithm (EGA) was the most consistent top performer. To formalize this, we ranked each of the six methods from 1 (best) to 6 (worst) in each scenario. The average ranks were: \textbf{EGA} ($1.58$), \textbf{GA} ($1.67$), \textbf{PSO} ($2.58$), \textbf{BSP} ($4.58$), \textbf{BSP-EW} ($4.58$), and \textbf{COP} ($4.75$).

This analysis shows that while GA achieved more outright victories, \textbf{EGA was more consistently ranked first or second}, making it arguably the most robust and reliable algorithm overall. The baseline methods, particularly COP and BSP-EW, consistently lagged, demonstrating the inadequacy of simple, non-adaptive heuristics in complex stochastic environments.

\subsection{Analysis of Computational Time}

The superior performance of the hyper-heuristics comes with a clear trade-off in computational time. The methods can be distinctly categorized into two groups:

\begin{itemize}
    \item \textbf{Fast, Low-Performance Methods:} The \textbf{BSP} and \textbf{COP} policies are exceptionally fast, with training times typically under two minutes and negligible evaluation time. This speed makes them suitable for generating quick baseline policies, but their poor rewards show that this comes at a steep cost in solution quality. Is worth noting that we use Monte Carlo optimization for parameter tuning; a more exhaustive search could improve their performance, albeit at a higher computational cost.

    \item \textbf{Computationally Intensive, High-Performance Methods:} \textbf{EGA, GA, and PSO} all require a significant computational investment for training, often ranging from minutes to several hours. Among these, there is no single fastest algorithm; their training times vary significantly depending on the scenario. For instance, in \texttt{I-4-2-30-H-Demand}, GA ($675.54s$) was notably faster than EGA ($1093.72s$), while in other large-scale scenarios their times were more comparable.
\end{itemize}

The key takeaway is that the one-time, offline training cost of a hyper-heuristic, such as GA or EGA, consistently yields a superior policy. Since evaluation times for all methods are very low (typically under 35 seconds even for the most complex scenarios), a well-trained policy can be deployed efficiently in a real-world setting.

\subsection{Scenarios of Hyper-heuristic Superiority}

While GA and EGA performed well across the board, their margin of victory was particularly pronounced in scenarios characterized by high complexity, uncertainty, and long-term planning horizons.

\begin{itemize}
    \item \textbf{High-Uncertainty Environments (Group 2):} In scenarios with high demand or supply volatility (Group 2), the performance gap between the hyper-heuristics and simpler methods like COP widened dramatically. For example, in \texttt{I-10-4-30-L}, GA's total reward ($-7132.67$) represented a \textbf{$14.21\%$ improvement} over the BSP baseline, whereas COP's performance was over $13\%$ worse ($-13.14\%$). This provides strong evidence of the value of sophisticated algorithms in turbulent environments where risk management is key.
    \item \textbf{Long-Horizon and Large-Scale Scenarios:} The advantage of the top-tier algorithms was amplified over longer time horizons and with more items. In \texttt{I-10-4-60-H} (long horizon), GA delivered a \textbf{$9.49\%$ improvement} over the baseline. In the most complex instance, \texttt{I-20-5-60-H}, EGA achieved a \textbf{$9.07\%$ improvement}. This suggests that the small, compounding errors of suboptimal policies become highly detrimental over time, making a robust policy from GA or EGA indispensable for strategic, long-term planning.
\end{itemize}

\subsection{Strategic Insights from Cost Metrics (Wastage, Lost Sales, Holding)}

Analyzing the individual cost components reveals the underlying strategies of the different algorithms.

\begin{itemize}
    \item \textbf{Minimizing Lost Sales (Prioritizing Customer Service):} GA and EGA consistently excelled at minimizing lost sales costs. They achieve this by learning to maintain a robust inventory buffer, ensuring high service levels, which directly translates to a higher total reward (lower total cost). This characteristic makes them the ideal choice for businesses where customer retention and market share are crucial.

    \item \textbf{Minimizing Inventory Costs (Prioritizing Lean Operations):} If the primary goal were to minimize holding costs, \textbf{COP} would appear effective due to its consistently low average inventory levels. However, this "lean" approach comes with a trade-off; it proves to be an extremely risky strategy, evidenced by high lost sales costs in nearly every complex scenario. Its simplicity, however, offers transparency, which might be preferred in situations where managers value the ability to manually fine-tune an intuitive policy over achieving optimal performance.

    \item \textbf{Wastage Cost: A Proxy for Policy Inefficiency:} A deeper look reveals that wastage is a better indicator of policy inefficiency than it is of high inventory. A counter-intuitive but crucial finding is that the supposedly "lean" \textbf{COP} algorithm often incurs the \textbf{highest wastage costs}, particularly in large-scale, high-uncertainty scenarios (e.g., in \texttt{I-20-5-60-H}, COP's wastage is $73.37$ vs. EGA's $10.37$). This occurs because simple, \textit{reactive} policies like COP create a "bullwhip effect," placing infrequent, large "panic orders" that risk expiry if not immediately consumed. In contrast, GA and EGA learn \textit{proactive}, smooth ordering policies. Furthermore, hyper-heuristics can learn sophisticated multi-supplier sourcing strategies---using cheap, high minimum order quantity suppliers for baseline demand and flexible, premium suppliers for demand spikes---a nuance that simple heuristics cannot capture, leading directly to their higher wastage.

    \item \textbf{The Balanced and Superior Approach:} The hyper-heuristic methods represent the most effective balance of these competing objectives. It consistently finds policies that minimize costly lost sales without creating excessive wastage or holding costs, demonstrating a sophisticated and holistic approach that leads to their superior overall performance.
\end{itemize}

In summary, we suggest \textbf{EGA} or \textbf{GA} hyper-heuristic population methods for profit maximization and ensuring high customer service, accepting moderately higher holding costs as the price for robustness. We indicate COP or BSP only if the explicit business strategy is to minimize inventory on-hand at all costs, accepting frequent and severe stockouts as a consequence.

\begin{figure}[H]
    \centering
    \includegraphics[width=\textwidth]{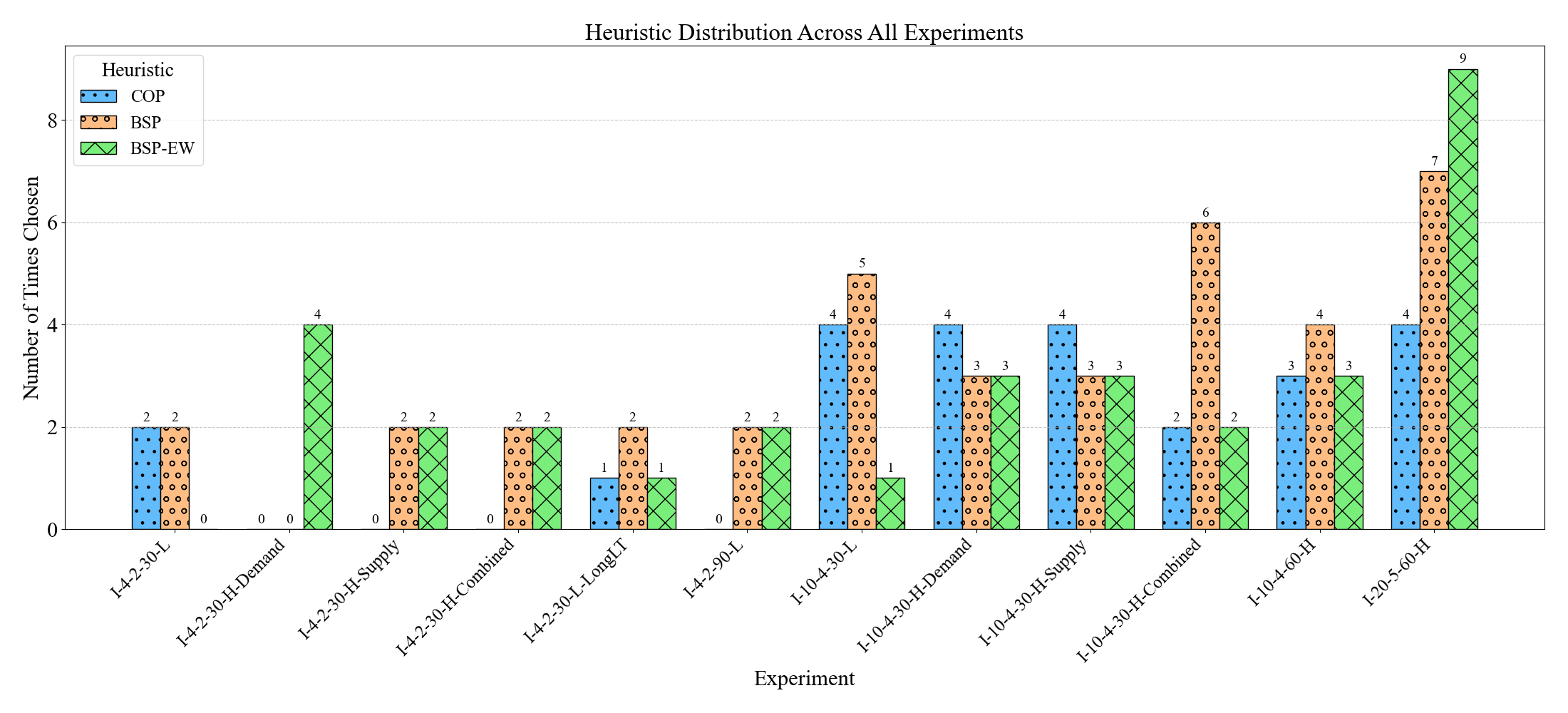}
    \caption{Distribution of baseline heuristic components (COP, BSP, BSP-EW) selected by the hyper-heuristic using EGA across all experimental scenarios. Each bar group represents one scenario, showing which heuristic was incorporated into the best-found hybrid policy.}
    \label{fig:heuristic_distribution}
\end{figure}

Further insight into the strategic choices made by the hyper-heuristics can be gleaned by examining the underlying heuristic components selected by EGA, the best hyper-heuristic, to build its hybrid policies, as shown in Figure \ref{fig:heuristic_distribution}.

Based on the provided experimental results, Figure \ref{fig:heuristic_distribution} illustrates which baseline heuristic (COP, BSP, or BSP-EW) was most frequently incorporated into the final, best-performing hybrid policies discovered by the hyper-heuristics using EGA. 
A pattern emerges: the Expected Wastage-Aware Base Stock Policy (BSP-EW) demonstrates a growing influence in the largest, most complex, and highest-uncertainty scenarios, particularly \texttt{I-10-4-60-H} and the extreme-scale \texttt{I-20-5-60-H}.

This trend is likely driven by several factors. 
First, as the number of items and the planning horizon increase, the financial risk from accumulated wastage becomes a critical cost driver. 
This makes the explicit, forward-looking wastage calculation within BSP-EW an indispensable logical component for the hyper-heuristic to leverage. 
Second, in these high-dimensionality environments with a low signal-to-noise ratio, the nuanced logic of BSP-EW---which tempers ordering decisions based on the age of existing stock---provides a superior guiding mechanism compared to the bluntness of a fixed order quantity (COP) or a simple reactive policy (BSP). 
The hyper-heuristics effectively learn that in the most challenging environments, embedding a sophisticated, wastage-aware logic is key to navigating the trade-offs and achieving a high-performing policy, even if the standalone BSP-EW policy performs poorly on its own.

\section{Deeper Insights and Strategic Implications}

Beyond the primary conclusions, a closer examination of the results offers further insights into the architectural advantages of the algorithms and the strategic trade-offs inherent in inventory management.

\subsection{Performance of Runner-Up Hyper-heuristics}

While GA and EGA emerged as the top-performing hyper-heuristics, PSO proved to be a highly competitive alternative. It consistently outperformed the baselines, achieving a notable average rank of 2.92. In certain instances, such as the \texttt{I-4-2-30-H-Combined} scenario, the performance of the leading methods converged. Here, EGA's reward ($-2902.81$) was comparable to that of GA ($-2768.23$), which PSO matched identically. This convergence suggests that specific search space structures are better suited to the distinct strategies employed by different hyper-heuristics, thereby reinforcing the importance of selecting an appropriate search method.

\subsection{Scalability and Performance Degradation}

The results clearly show how computational time for hyper-heuristics scales significantly with problem complexity. The training time for EGA, for example, increased by over \textbf{$50\times$} from one of the simplest scenarios ($119.65s$ for \texttt{I-4-2-30-L}) to the most complex ($6530.47s$ for \texttt{I-20-5-60-H}). This scalability is a critical consideration for practical implementation. Notably, our experiments did not employ parallelization, ensuring a fair comparison against the baseline heuristics. However, these population-based methods are highly amenable to parallel processing; for example, the fitness evaluations for a population can be easily distributed across multiple threads.

This scaling challenge suggests that all algorithms struggle with the "curse of dimensionality." The sheer size of the decision space in scenarios like \texttt{I-20-5-60-H} makes it incredibly difficult for any algorithm to find a truly global optimum in a reasonable time. The policies found are likely high-quality local optima, and at this scale, the choice of algorithm becomes a trade-off between potentially marginal performance gains and significant additional computational investment.

%\newpage

\section{Conclusion}

This work presented a comprehensive evaluation of multiple algorithms for solving complex, stochastic multi-item, multi-supplier perishable item inventory optimization problems. Through a series of twelve progressively challenging experimental scenarios, we compared the performance of simple heuristics (BSP, COP), enhanced baselines (BSP-EW), and advanced hyper-heuristics (GA, PSO, EGA).

Our primary finding is the consistent and robust superiority of the \textbf{hyper-heuristic approach}. While no single algorithm won every scenario, the results show a clear hierarchy of performance. 
The \textbf{Genetic Algorithm (GA)} secured the most individual victories, while the \textbf{Enhanced Genetic Algorithm (EGA)} proved to be the most consistent performer with the best average rank across all experiments. 
This work highlights the substantial performance improvement that advanced search methods provide in environments that force difficult trade-offs, such as minimizing waste for perishable goods. The ability of top-tier hyper-heuristics, such as GA and EGA, to select the appropriate heuristic for each item, balance inventory levels, and optimize supplier choice and order timing is crucial for minimizing costs and maximizing service levels in such an uncertain domain.

While simple heuristics provide rapid solutions, their performance degrades significantly under uncertainty, limiting their utility to that of a performance baseline. To ensure a fair comparison, both the baseline heuristics and the hyper-heuristics operated within the same policy space. Furthermore, all evaluated hyper-heuristic search methods---GA, PSO, and EGA---were allocated identical computational budgets in terms of population size and number of generations. Although \textbf{hyper-heuristics require a greater upfront training investment}, this one-time computational cost is justified by the superior quality and long-term stability of the resulting policy. Consequently, \textbf{the choice between GA and EGA offers the most favorable trade-off}, with GA being the preference for achieving potentially the highest reward and EGA being the recommended algorithm where robust, consistently high performance is the primary goal.

Additionally, our analysis highlighted the practical limits of optimization under extreme conditions. In the largest and most uncertain scenario, the performance gap between the top hyper-heuristics (EGA, GA, and PSO) narrowed, suggesting that the "curse of dimensionality" and an overwhelmingly low signal-to-noise ratio begin to act as equalizers. At this scale, the problem's inherent chaos may limit the maximum achievable performance, making factors such as convergence speed and computational cost increasingly important in selecting an algorithm.
\subsection{Future Work}

Building on these findings, future research could explore several promising avenues, such as hybridizing algorithms like PSO and EGA to balance exploration and fine-tuning for speed and performance at extreme scales. Further work could also incorporate dynamic cost structures or non-stationary demand patterns to test algorithmic adaptability and enhance policy robustness. An equally interesting direction is the development of policies that can find approximated policies that can take advantage of additional degrees of freedom, but without sacrificing too much when having to look into a larger search space. RL or Approximated Dynamic Programming might be an adequate method for that. Finally, extending the problem to a multi-echelon supply chain network would offer a more holistic view of inventory flows and present a greater challenge for these optimization techniques.

\section*{Acknowledgments}
The authors are also grateful for the financial support provided by CNPq (403735/2021-1; 309385/2021-0) and FAPESP (2013/07375-0; 2022/05803-3).

%%%%%%%%%%%%%%%%%%%%%%%%%%%%%%%%%%%%%%%%%%%%%%%%

\bibliographystyle{plain}
%\bibliographystyle{elsarticle-harv} 

%\section*{\refname}
\bibliography{mybibfile}

\end{document}